\documentclass[conference]{IEEEtran}

\usepackage{amsmath,amssymb}
\usepackage{graphicx}
\usepackage{booktabs}
\usepackage{url}
\usepackage{hyperref}
\usepackage{xcolor}
\usepackage{cite}
\usepackage{balance}
\usepackage{microtype}
\usepackage{placeins}
\usepackage{mdframed}
\usepackage{tikz}
\usetikzlibrary{arrows.meta}

\graphicspath{{./}{./figures/}{./results/classifiers/}}
\hypersetup{hidelinks}
\IfFileExists{mathptmx.sty}{\usepackage{mathptmx}}{}

\begin{document}

\title{A Generalizable and Explainable Framework for Synthetic Video Detection Using First-Digit Gradient Statistics}

\author{
    \IEEEauthorblockN{Sidharth Shanu}
    \IEEEauthorblockA{\textit{B.Tech CSE}\\
        IIT Jodhpur, India\\
        sidharthshanukt@gmail.com}
    \and
    \IEEEauthorblockN{Gautam Kumar}
    \IEEEauthorblockA{\textit{School of Automation \& Robotics}\\
        GGSIPU, New Delhi, India\\
        gautamkumar.kt@gmail.com}
    \and
    \IEEEauthorblockN{Tej Singh}
    \IEEEauthorblockA{\textit{Centre for Artificial Intelligence}\\
        MITS Gwalior, India\\
        tejs@mitsgwalior.in}
}

\maketitle

\setcounter{topnumber}{9}
\setcounter{bottomnumber}{9}
\setcounter{totalnumber}{9}
\renewcommand{\topfraction}{0.99}
\renewcommand{\bottomfraction}{0.99}
\renewcommand{\textfraction}{0.01}
\renewcommand{\floatpagefraction}{0.5}
\raggedbottom

\begin{abstract}
Ai video generators have not only become harder to detect but are used to generate
a diverse set of scenarios from landscapes to street view to animal videos. This 
creates a problem where cnn based detectors are effective but offer no insight in
the inner working while the forensics based detectors are often pretrained for a 
set scenario or become too complex to derive meaningful insights. We present a 
unique way of ai video detection using sobel gradient values analysed using first 
digit law. Using Linear disriminant analysis, we visualize the discrimantory signal 
while mlp is used for classification. The detection method has no generator and scene 
specific features. We have ensured that model has no knowledge of container formats, 
codec,bitrate or compression artifact. The model is trained and tested on GenBuster-200K~\cite{wen2025busterxmllmpoweredaigeneratedvideo},
GenBusterBench~\cite{wen2025busterxmllmpoweredaigeneratedvideo},
GeneVA~\cite{kang2025geneva}, FaceForensics++ C23~\cite{rossler2019faceforensics},
and CelebDF~\cite{li2020celebdf}.
We also show how zero shot detection fails even though the feature set has discrimintory signal.

\end{abstract}


\section{Introduction}

The cost of producing an AI-generated video has shrunk rapidly in recent
years, driven by advances in generative modelling~\cite{goodfellow2014gan,
ho2020ddpm,ho2022video} and increasingly efficient model
designs~\cite{blattmann2023svd}, moving synthetic video from a research
curiosity to a regular part of everyday online feeds.  As the technology has
matured, the obvious artefacts that a human viewer could once identify have
steadily diminished, making the output ever more photorealistic and creating
a pressing need for automated detection mechanisms and new human-interpretable
signals~\cite{wang2020cnn,gragnaniello2021easy}.  Detection and generation
thus form a cat-and-mouse game, but an asymmetric one: any published
detection artefact can be adopted as a training signal by the next generation
of models, hardening them against that artefact, whereas detectors gain no
comparable advantage from advances in generation.  This asymmetry demands
that sustained effort be invested in finding and refining new detection
artefacts.

Existing detection mechanisms can be categorised into two broad families:
forensics-based and heavy machine-learning-based.  In the forensic approach
the goal is to first identify a discriminative signal and then train a
low-capacity model to discern on that
signal~\cite{marra2018social,marra2019fingerprints,frank2020leveraging}.
Because a human decision determines the signal of choice, this approach keeps
the detector interpretable and isolates which signals are actually useful.
Heavy machine-learning models---CNNs, MLLMs, and other large
architectures---instead take raw pixels with little or no hand-designed
preprocessing of the input, and learn, refine, and construct the decision
boundary end-to-end from
data~\cite{rossler2019faceforensics,wang2020cnn,wen2025busterxmllmpoweredaigeneratedvideo,wen2025busterxunifiedcrossmodalaigenerated}.
Although this strategy has proven very powerful, it offers limited insight
into what has been learned, so that insight cannot be transferred to bootstrap
the next model, and each new detector instead requires large amounts of
training data and heavy computation~\cite{gragnaniello2021easy}.

In forensics-based detection, the sole goal of extracting a discernible signal
requires compacting the information contained in each video or file; failing
to do so causes the model to lean more heavily on the machine-learning
component and forfeits the benefit of forensic analysis.  First-digit signal
is a useful tool in this setting because it substantially reduces the amount
of information while also making the process more intuitive.  First-digit
laws --- primarily Benford's law --- have previously been used for
AI-generated content detection, but mainly via DCT and other frequency
transforms, because those coefficients are known to follow Benford's
law~\cite{bonettini2020benford,frank2020leveraging,chen2009firstdigit}.  In
this paper, we instead search for a discriminatory signal using the Sobel
operator, a discrete differentiation operator that belongs to a completely
different class of transform.

Existing mathematical theory does not predict that Sobel-gradient magnitudes
should follow Benford's law, and and the present frame-level and temporal
Sobel analysis (Section~\ref{subsec:benford}) confirm that they do not.  Even
though direct goodness-of-fit comparisons against the ideal Benford
distribution fail, we find that the MLP classifier is able to construct a
clean decision boundary, achieving over 90\% accuracy and AUC on
AI-generated video detection, with LDA confirming strong class separability
along a low-dimensional discriminant axis.  The key takeaway is that even
though the first-digit statistics do not follow Benford's law, they still
carry a clear and discernible discriminatory signal between AI-generated and
real videos.

The paper is organised as follows.  Section~\ref{sec:related} reviews related
work.  Section~\ref{sec:methodology} describes the methodology.
Section~\ref{sec:experiments} details the experimental setup.
Section~\ref{sec:results} reports results across five datasets.
Section~\ref{sec:conclusion} concludes.

\section{Related Work}
\label{sec:related}

\textbf{Generative models.}
Goodfellow et al.~\cite{goodfellow2014gan} introduced generative adversarial networks
(GANs), enabling photorealistic image synthesis at scale.  Ho et al.~\cite{ho2020ddpm}
subsequently showed that denoising diffusion probabilistic models surpass GANs in sample
quality and training stability.  Both architectures, together with transformer-based video
generators, now underpin a wave of synthetic media that undermines the integrity of visual
evidence across news, social media, and legal proceedings.

\textbf{CNN-based synthetic image and video detection.}
Wang et al.~\cite{wang2020cnn} demonstrated that a ResNet classifier trained on
ProGAN outputs transfers surprisingly well to other GAN architectures, establishing the
template for CNN-based generalisation.  Gragnaniello et al.~\cite{gragnaniello2021easy}
provided a critical analysis of this family, showing that generalisation degrades
substantially when generator architecture or post-processing differs from the training
distribution.  For video, FaceForensics++~\cite{rossler2019faceforensics} benchmarked
neural detectors across face-swap and expression-reenactment forgeries under varying
compression; Celeb-DF~\cite{li2020celebdf} revealed that detectors trained on one
forgery pipeline generalise poorly to another.  All these methods require GPU hardware
for training and inference and offer no interpretable account of which image or video
properties drive classification.

\textbf{Frequency-domain and spectral fingerprints.}
Frank et al.~\cite{frank2020leveraging} showed that GAN-generated images exhibit
distinctive peaks in the discrete cosine transform (DCT) spectrum arising from periodic
upsampling operations, enabling high classification accuracy.  Marra et
al.~\cite{marra2018social,marra2019fingerprints} and Yu et al.~\cite{yu2019attributing}
extended this to GAN-fingerprint attribution.  However, these spectral signals live in
high-frequency DCT coefficients---precisely the components discarded by JPEG recompression
applied universally by social-media platforms---limiting applicability to unprocessed
content.

\textbf{First-digit laws in image and video forensics.}
Benford's law predicts that the leading significant digit of naturally occurring
numerical sequences follows a logarithmic distribution.  Bonettini et
al.~\cite{bonettini2020benford} applied this principle to DCT coefficient distributions
of still images, measuring deviations between real and GAN-generated samples; the
mechanism shares the compression vulnerability of spectral fingerprints because the
signal resides in high-frequency coefficients.  Chen and Shi~\cite{chen2009firstdigit}
earlier applied first-digit statistics to quantised DCT AC coefficients in MPEG video
streams to detect double-compression artefacts, establishing that the first-digit law
carries forensically useful information in the video domain.  

\textbf{Video-specific AI-generated content detection.}
BusterX~\cite{wen2025busterxmllmpoweredaigeneratedvideo} and its successor
BusterX++~\cite{wen2025busterxunifiedcrossmodalaigenerated} approach AI-generated video detection through large
multimodal language models, requiring substantial GPU compute and offering limited
interpretability of the detection criterion.  GeneVA~\cite{kang2025geneva}
provides human-annotated artefact labels for text-to-video generative models,
characterising the qualitative failure modes of current generators and motivating
quantitative forensic features that must be generalisable across generators and robust
to post-processing.  The present work addresses this need by creating a temporal-derivative feature bank
built from a 3-D Sobel gradient is added alongside the spatial bank, encoding
frame-to-frame pixel change rates that synthetic generators produce inconsistently
relative to genuine optical flow.

\section{Methodology}
\label{sec:methodology}

The pipeline consists of two complementary feature banks --- \emph{spatial} and
\emph{temporal-derivative} --- extracted from a fixed set of ten image channels.
Both banks share the same gradient and first-digit analysis machinery; their
concatenation yields a 800-dimensional fixed-length vector that is the sole input
to the classifier.

\begin{figure*}[!t]
\centering
\includegraphics[width=0.74\textwidth]{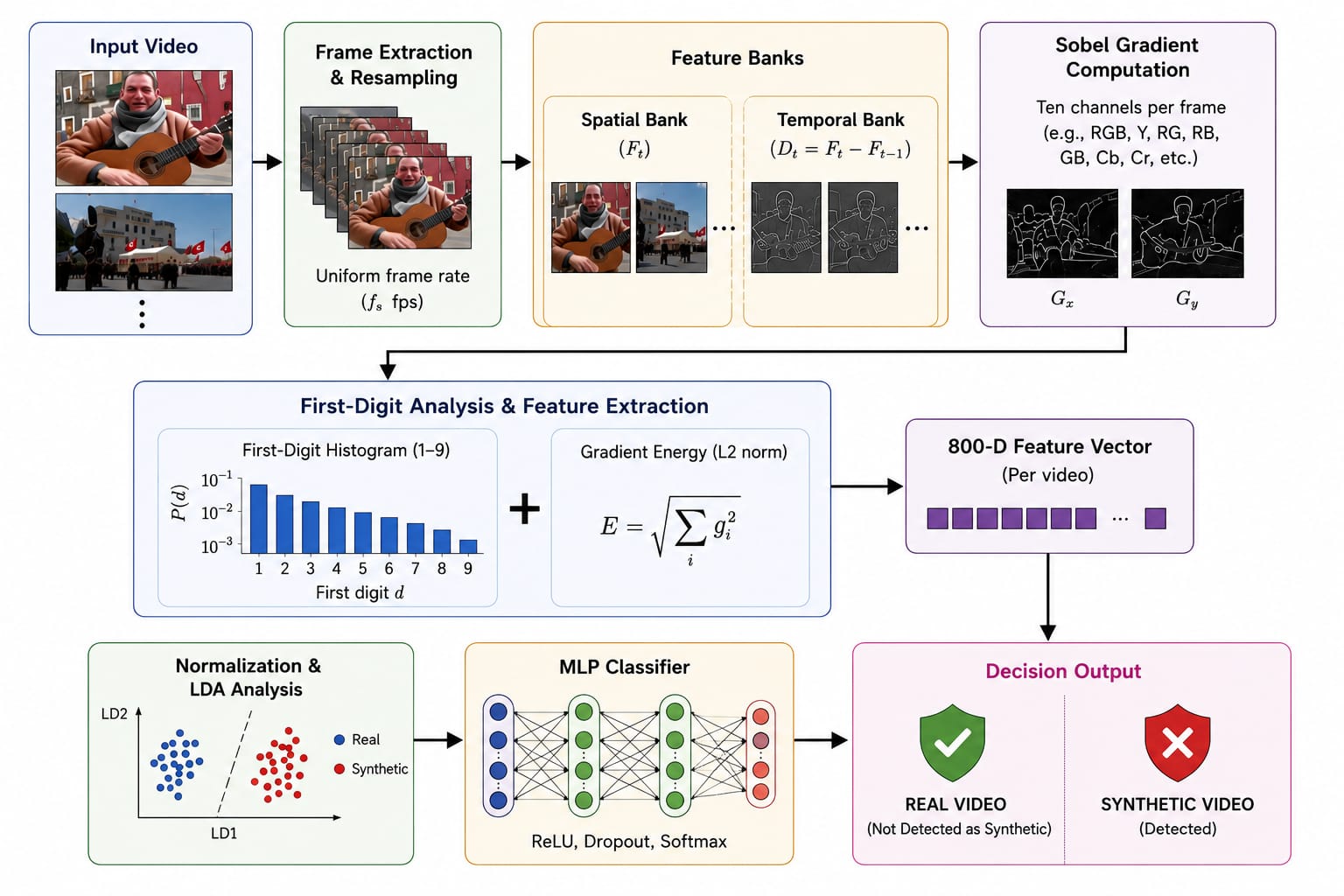}
\caption{A Generalizable and Explainable Framework for Synthetic Video
Detection Using First-Digit Gradient Statistics.  Each video is resampled to
a uniform frame rate, split into spatial ($F_t$) and temporal-derivative
($D_t$) banks, and passed through Sobel gradient computation over ten
per-frame channels.  The resulting gradient responses are summarised as
first-digit histograms and energy terms, concatenated into an 800-D feature
vector, and either projected with LDA for visualisation or classified
directly by the MLP to yield a real/synthetic decision.}
\label{fig:pipeline}
\end{figure*}

\textbf{Channel set.}
Ten channels are derived from each BGR frame $F_t$ as shown in
Table~\ref{tab:channels}.  Luminance $\mathrm{br}$ provides a channel-agnostic
measure of structural edges.  The sharpness channel $\mathrm{sh} =
|\nabla^2\!\mathrm{br}|$ (Laplacian magnitude of luminance) encodes local
focus energy.  The signed channel-correlation differences $\mathrm{rg},\mathrm{rb},
\mathrm{gb}$ measure how tightly the three colour planes co-vary spatially.  The BT.601
chrominance channels $\mathrm{Cb},\mathrm{Cr}$ isolate colour from luminance.

\begin{table}[!t]
\centering
\caption{The ten feature channels derived from each frame.}
\label{tab:channels}
\renewcommand{\arraystretch}{1.15}
\setlength{\tabcolsep}{4pt}
\begin{tabular}{@{}lll@{}}
\toprule
\textbf{Symbol} & \textbf{Definition} & \textbf{Rationale} \\
\midrule
$r,g,b$      & Raw BGR colour planes & Per-channel edge structure \\
$\mathrm{br}$ & $0.299R{+}0.587G{+}0.114B$ & Luminance baseline \\
$\mathrm{sh}$ & $|\nabla^2\mathrm{br}|$ & Sharpness / focus energy \\
$\mathrm{rg}$ & $R-G$ & Channel-correlation check \\
$\mathrm{rb}$ & $R-B$ & Channel-correlation check \\
$\mathrm{gb}$ & $G-B$ & Channel-correlation check \\
$\mathrm{Cb}$ & $0.5{-}0.169R{-}0.331G{+}0.500B$ & BT.601 blue-difference chroma \\
$\mathrm{Cr}$ & $0.5{+}0.500R{-}0.419G{-}0.081B$ & BT.601 red-difference chroma \\
\bottomrule
\end{tabular}
\end{table}

\textbf{Frame selection.}
Videos in a real-world corpus are recorded at heterogeneous native frame
rates --- commonly 24, 25, or 30\,fps --- so processing frames at their
original rate would make the inter-frame motion depend on the source frame
rate rather than the scene content.  To normalise motion across all videos,
each video is first resampled to a common \emph{target frame rate}
$f_s$ (e.g.\ 4\,fps) by selecting the nearest native frame at uniform temporal
intervals of $1/f_s$ seconds, discarding the intervening frames.  At this rate, one
inter-frame interval corresponds to the same real-world time regardless of
whether the original video was shot at 24 or 30\,fps.  The target sampling
rate must satisfy
\begin{equation}
f_s \;\le\; \mathrm{fps}_\mathrm{min},
\label{eq:fs_bound}
\end{equation}
where $\mathrm{fps}_\mathrm{min}$ denotes the lowest native frame rate
among all the videos in the dataset being processed.  Selecting an $f_s$
larger than this value would require sampling frames that are not present
in some of the lower-frame-rate videos, preventing a consistent resampling
across the dataset.

Spatial features (below) are computed independently on every resampled frame
$F_t$.  Temporal-derivative features are computed from the difference
between each consecutive pair of resampled frames,
\begin{equation}
D_t = F_{t} - F_{t-1},
\label{eq:tdiff}
\end{equation}
i.e.\ a simple first-order finite difference between adjacent resampled
frames, with no additional temporal filtering applied beforehand.  $F_t$
encodes \emph{what the scene looks like} at that instant, while $D_t$
encodes \emph{how each pixel is changing} between the two consecutive
sampled frames.

\textbf{Sobel gradient computation.}
For a single-channel image $I$, horizontal and vertical partial derivatives are
computed by convolving with a pair of $k{\times}k$ derivative kernels.  For $k=3$
these are the standard Sobel kernels:
\begin{equation}
K_x = \begin{pmatrix}-1&0&\phantom{-}1\\-2&0&\phantom{-}2\\-1&0&\phantom{-}1\end{pmatrix}\!,
\quad
K_y = \begin{pmatrix}-1&-2&-1\\\phantom{-}0&\phantom{-}0&\phantom{-}0\\\phantom{-}1&\phantom{-}2&\phantom{-}1\end{pmatrix}\!.
\label{eq:sobel_kernels}
\end{equation}
For $k=1$ the kernel reduces to a three-tap central difference.  Two kernel
sizes $k\in\{1,3\}$ and two directions $(\partial_x,\partial_y)$
are applied independently to each of the ten channels.

\emph{Spatial bank.}
The Sobel kernels are applied directly to each channel of the resampled
frame $F_t$, so the spatial response captures the distribution of edge
energies within a single image.

\emph{Temporal-derivative bank.}
The same Sobel kernels are applied to the corresponding channel of the
frame-difference image $D_t$ (Eq.~\ref{eq:tdiff}), so the temporal-derivative
response captures the distribution of edge energies in how the scene changes
between consecutive sampled frames.  The two banks therefore share identical
gradient machinery and differ only in which image --- $F_t$ or $D_t$ --- the
Sobel kernels are applied to.  Synthetic videos often present temporal
transitions that differ from those observed in naturally captured footage;
in our experiments these differences are frequently associated with a
shifted gradient-energy distribution in the temporal-derivative bank
relative to genuine videos.

\textbf{First-significant-digit histogram and energy.}
For each combination of bank, channel, kernel size and gradient direction,
the gradient response is spatially subsampled with a stride of eight pixels
in both dimensions to form the set $\mathcal{G}$.  The non-zero elements of
$\mathcal{G}$ are represented using a 9-bin first-significant-digit (FSD)
histogram, and a single root-mean-square energy is computed over all of
$\mathcal{G}$ (including any zero-valued entries), producing ten features
per group.
The leading digit of a positive response $g>0$ is
\begin{equation}
d(g) = \left\lfloor \frac{g}{10^{\lfloor\log_{10}g\rfloor}} \right\rfloor \in \{1,\ldots,9\},
\label{eq:leading_digit}
\end{equation}
and, writing $\mathcal{G}^{+}=\{g\in\mathcal{G}:g>0\}$ for the non-zero
subset, the histogram is
\begin{equation}
h[n] = \frac{1}{|\mathcal{G}^{+}|}\sum_{g\in\mathcal{G}^{+}}\mathbf{1}[d(g)=n], \quad n=1,\ldots,9,
\label{eq:fsd_hist}
\end{equation}
The histogram is L1-normalised so that $\sum_n h[n]=1$.
For natural data obeying Benford's law the expected bin
probability is $\log_{10}(1+1/n)$. Appended to the nine normalised
bins is the root-mean-square energy, taken over the full strided sample
$\mathcal{G}$ rather than $\mathcal{G}^{+}$ alone:
\begin{equation}
E = \sqrt{\frac{1}{|\mathcal{G}|}\sum_{g\in\mathcal{G}} g^2},
\label{eq:energy}
\end{equation}
The energy term gives a measure of the overall strength of the gradient
responses. This can help distinguish smoother or blurred synthesised areas,
which exhibit lower gradient magnitudes from the richer regions containing
stronger edge structure of a genuine face.

\textbf{Feature banks and final vector.}
Let $\mathcal{B}\in\{\mathrm{sp},\mathrm{td}\}$ denote the two banks.  For
the spatial bank, the per-group accumulator is updated for every resampled frame
and divided by the total number of resampled frames.  For the temporal-derivative bank,
the accumulator is updated for every consecutive resampled-frame pair
$(F_{t-1},F_t)$ and divided by the number of pairs.  Concatenating both
banks yields a 800-dimensional vector:
\begin{equation}
\mathbf{f} = \bigl[\mathbf{f}_\mathrm{td}^\top \;\Big|\; \mathbf{f}_\mathrm{sp}^\top\bigr]^\top \in \mathbb{R}^{800},
\label{eq:feat_vec}
\end{equation}
where $800 = 2\text{ banks}\times10\text{ channels}\times2\text{ kernels}\times
2\text{ directions}\times10\text{ values}$.  The vector is computed entirely
from deterministic, hand-crafted operations.

\section{Experimental Setup}
\label{sec:experiments}

\textbf{Hardware.}
All experiments---feature extraction, model training, and inference---are conducted on a
desktop workstation equipped with an Intel Core Ultra~7 270K Plus processor (3.70\,GHz),
32.0\,GB of installed RAM (31.6\,GB usable), and an NVIDIA GeForce RTX\,5070 GPU
(12\,GB VRAM), running a 64-bit operating system.  The feature extraction pipeline is
fully deterministic and hand-crafted; it applies no learned filters and requires no GPU
at any stage, so all feature computation runs on the CPU alone.  The GPU is used solely
for MLP training and inference.

\textbf{Feature extraction.}
The feature pipeline follows Section~\ref{sec:methodology} exactly.  Each video is
resampled to a common target frame rate and divided into non-overlapping temporal windows
of $T$ frames; spatial and temporal-derivative Sobel responses are accumulated across all
ten channels at two kernel sizes and two gradient directions, yielding one fixed
800-dimensional vector per video.  No data augmentation is applied at the feature
extraction stage.

\textbf{Classifier and training.}
A three-layer multi-layer perceptron (MLP) with ReLU activations and dropout
regularisation is trained on the 800-dimensional feature vectors. Class imbalance is
addressed with focal loss~\cite{lin2017focal}.

\textbf{LDA and discriminability analysis.}
Linear discriminant analysis (LDA) is applied to the 800-dimensional feature space
to visualise class separability and to quantify the ratio of between-class to
within-class variance.  Per-feature discriminability is measured by Cohen's~$d$,
computed as the standardised mean difference between the real and synthetic video
feature distributions.  

\textbf{Datasets and evaluation protocol.}
Five datasets are used.
\textbf{GenBuster-200K}~\cite{wen2025busterxmllmpoweredaigeneratedvideo} contains 14,000 videos: 6,918 real clips and
7,082 generated by twelve text-to-video models (CogVideoX, EasyAnimate,
Gen3, HunyuanVideo, JiMeng, Kling, LTXVideo, Luma, Pika, Sora, ViDu,
WanX); a stratified 60/10/30 split is used for train, validation, and test.
\textbf{GenBusterBench}~\cite{wen2025busterxmllmpoweredaigeneratedvideo} provides 3,150 videos (1,000 real; 2,150 from 23
generators) evaluated across three benchmark-defined subsets: a
\emph{standard} subset sharing generator families with GenBuster-200K, an
\emph{extended} subset of additional contemporary generators, and a
\emph{wild} subset of the latest-generation models released after the
benchmark was compiled.
\textbf{GenVA}~\cite{kang2025geneva} contains 16,351 synthetic videos from
three generators (Pika~v1, Sora, VideoCrafter-2); it carries no real-video
split and is used as a zero-shot cross-domain test, with real videos drawn
from the GenBuster-200K test split.
\textbf{FaceForensics++}~\cite{rossler2019faceforensics} (C23 compression)
contains 1,000 videos per class across five forgery methods (Deepfakes,
Face2Face, FaceShifter, FaceSwap, NeuralTextures) and 1,000 originals; a
1,600-dimensional face+background feature is used (two concatenated 800-D
vectors: one for the detected face region and one for the zeroed-face
background frame).
\textbf{CelebDF}~\cite{li2020celebdf} contains 6,528 videos (889 real, drawn
from Celeb-real and YouTube-real; 5,639 Celeb-synthesis face-swap fakes); an
80/20 stratified split (seed 42) is used, with the standard 800-D video
feature and no face-crop bank, to check whether the detector generalises to a
second, independently sourced face-swap benchmark beyond FaceForensics++.
Primary metrics are ROC-AUC and balanced accuracy (BalAcc); per-class recall
is reported for multi-class experiments.

\begin{figure}[!ht]
\centering
\includegraphics[width=0.72\columnwidth]{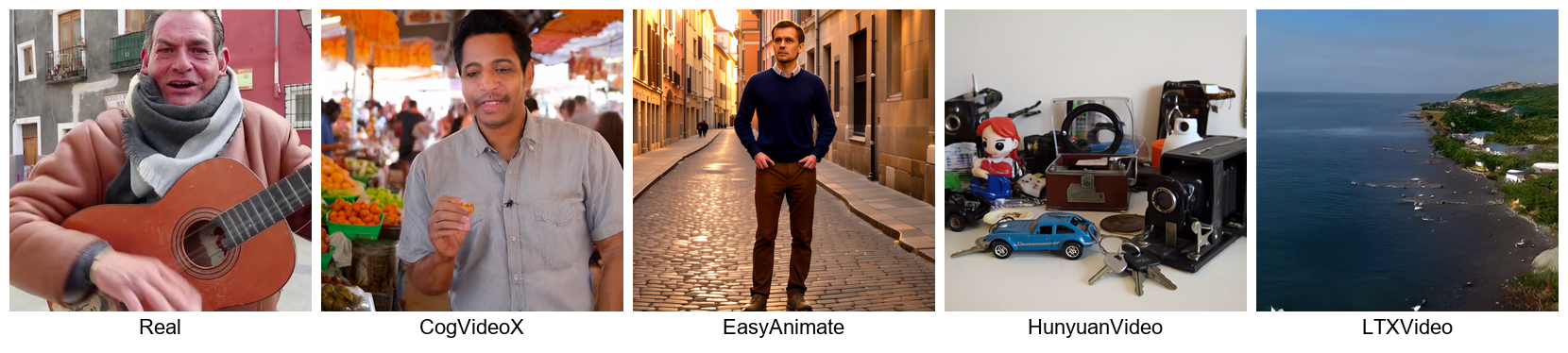}
\caption{Representative frames from GenBuster-200K.  From left to right: real,
CogVideoX, EasyAnimate, HunyuanVideo, LTXVideo.}
\label{fig:gb200k_samples}
\end{figure}

\begin{figure}[!ht]
\centering
\includegraphics[width=0.56\columnwidth]{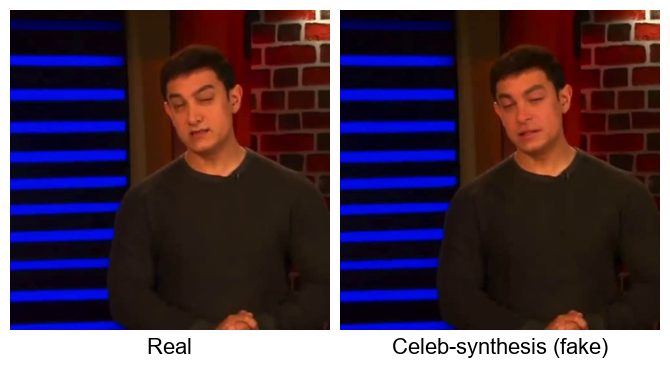}
\caption{Representative frame pair from CelebDF.  Left: real (Celeb-real).
Right: Celeb-synthesis face-swap fake of the same subject.  The manipulation
is visually near-imperceptible, consistent with the low 0.624 AUC reported in
Section~\ref{subsec:celebdf}.}
\label{fig:celebdf_samples}
\end{figure}

\section{Results}
\label{sec:results}

% ── A. GenBuster-200K ─────────────────────────────────────────────────────────
\subsection{GenBuster-200K: In-Domain Binary Detection}

Figure~\ref{fig:gb200k_samples} shows representative real and generated frames
from GenBuster-200K.

\begin{figure}[htbp]
\centering
\includegraphics[width=0.72\columnwidth]{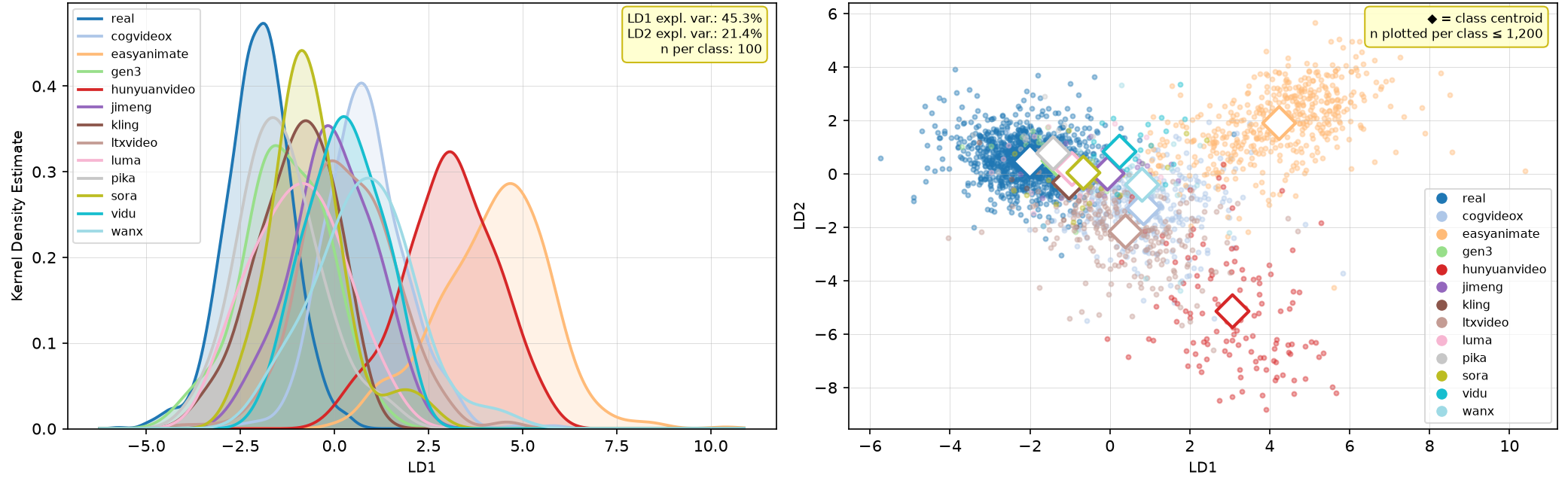}
\caption{LDA on the GenBuster-200K test split.  Left: LD1 kernel-density
estimate per generator.  Right: LD1\,vs.\,LD2 scatter with class centroids
($\blacklozenge$).}
\label{fig:lda_gb200k}
\end{figure}

\begin{figure}[!ht]
\centering
\includegraphics[width=0.3\columnwidth]{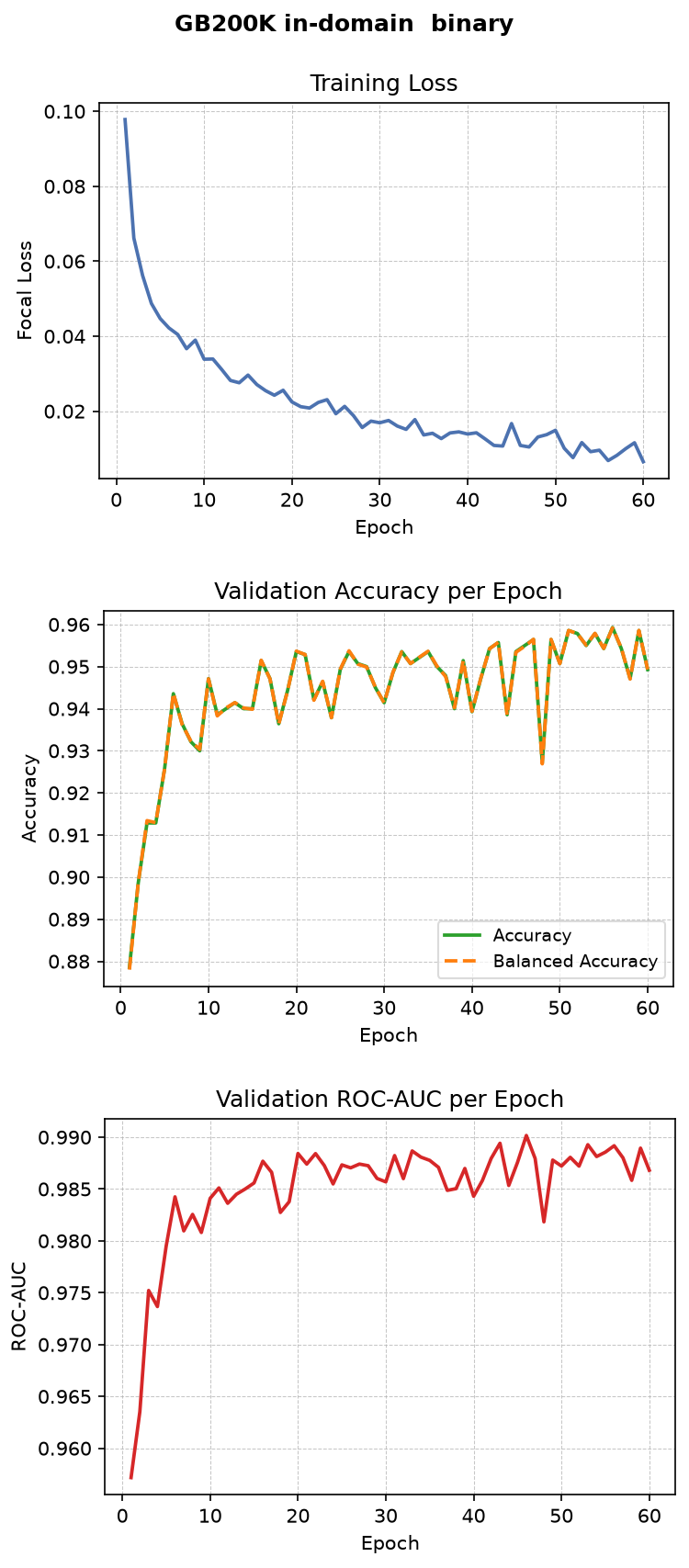}
\caption{GenBuster-200K MLP training curves (loss, balanced accuracy, and
ROC-AUC per epoch on the validation split).}
\label{fig:gb200k_curves}
\end{figure}

Fig.~\ref{fig:lda_gb200k} shows the LDA projection on the GenBuster-200K
test split.  The first discriminant cleanly separates the twelve AI
generators from the real class. The three-layer MLP trained on the 60\%
training split achieves ROC-AUC\,=\,0.986 and balanced accuracy 0.945
(Table~\ref{tab:gb200k}); training convergence is shown in
Fig.~\ref{fig:gb200k_curves}.

Training was performed on 14,000 videos rather than the full
$\sim$200,000-video corpus.  The grouped LDA in
Fig.~\ref{fig:lda_gb200k_grouped} reveals why additional volume from the
same generators yields diminishing returns: each generator family occupies a
distinct direction in the discriminant space, and that direction is
well-represented after a few hundred clips.  The generalisation bottleneck is
\emph{directional diversity} across generator families, not sample quantity
within any one family.  A model trained on 200\,K clips of the same twelve
generators would learn the same directions and produce essentially the same
decision boundary while incurring an order of magnitude more compute and
creating a false impression of data richness without improving
out-of-distribution detection.

\begin{figure}[!h]
\centering
\includegraphics[width=0.62\columnwidth]{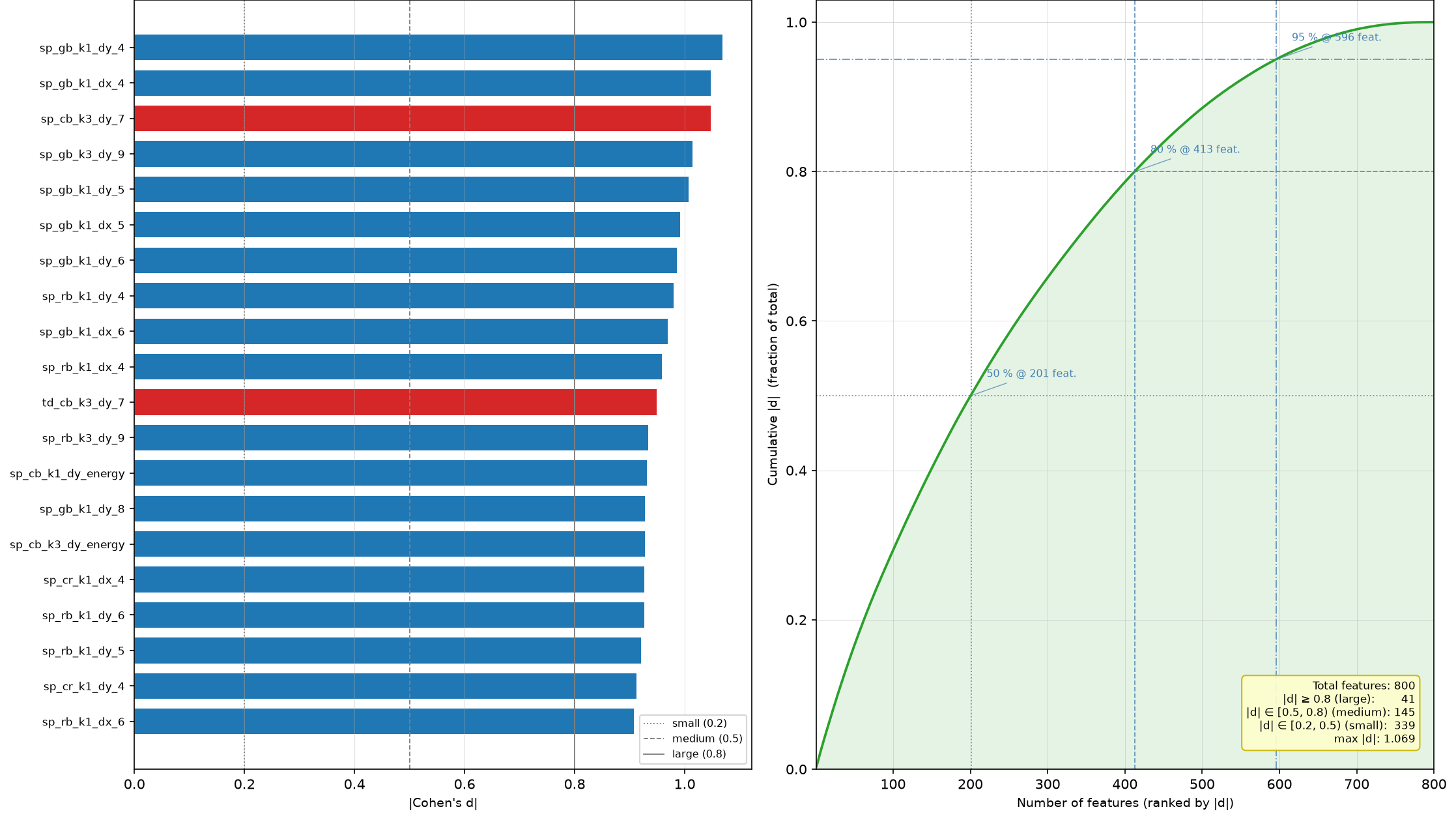}
\caption{Cohen's $d$ on GenBuster-200K.  Left: top-20 features by $|d|$
(blue = synthetic$>$real, red = synthetic$<$real).  Right: cumulative
discriminative power; crosshairs at 50\%, 80\%, 95\%.}
\label{fig:cohen_gb200k}
\end{figure}

Fig.~\ref{fig:cohen_gb200k} shows the per-feature Cohen's $d$ analysis.
The top discriminative features are spatial-channel gradient bins
(\texttt{sp\_ch\_*}), with 41 features reaching a large effect size
($|d|\geq0.8$).  The cumulative power curve reaches 50\% of total
discriminative power at approximately 41 features and 80\% at
${\sim}145$ features, indicating that the signal is moderately
concentrated but not dominated by a handful of features---consistent
with a broadly distributed artefact signature spanning multiple channels
and gradient directions. Synthetic videos systematically exhibit lower
gradient energy when compared to the real footage, as indicated by the
predominance of red bars (fake $<$ real) across the top features.
Generative models tend to produce smoother pixel transitions on average, 
and a corresponding shift can be observed in the first digit histogram
relative to those of naturally captured, edge rich scenes.

Table~\ref{tab:gb200k_pergen} breaks the in-domain result down by generator.
Generators that produce larger and more consistent gradient deviations, such
as EasyAnimate (99.6\%), HunyuanVideo (100\%), and ViDu (97.7\%), achieve high
detection rates, whereas those whose gradient-feature distributions lie
closer to real videos, such as Gen3 (50.0\%) and Kling (64.7\%), are harder
to distinguish from real footage at the chosen threshold even though they
were included in the training set.

\begin{table}[!t]
\centering
\caption{Per-generator detection on the GenBuster-200K in-domain test set.}
\label{tab:gb200k_pergen}
\renewcommand{\arraystretch}{1.12}
\setlength{\tabcolsep}{4pt}
\begin{tabular}{@{}lrcc@{}}
\toprule
\textbf{Generator} & $n_\text{test}$ & \textbf{Det.\ rate} & \textbf{AUC} \\
\midrule
CogVideoX    & 534 & 93.1\% & 0.988 \\
EasyAnimate  & 733 & 99.6\% & 0.998 \\
Gen3         &  32 & 50.0\% & 0.922 \\
HunyuanVideo & 202 &100.0\% & 1.000 \\
JiMeng       &  31 & 80.6\% & 0.965 \\
Kling        &  34 & 64.7\% & 0.954 \\
LTXVideo     & 363 & 93.1\% & 0.985 \\
Luma         &  30 & 73.3\% & 0.960 \\
Pika         &  23 & 73.9\% & 0.969 \\
Sora         &  59 & 74.6\% & 0.955 \\
ViDu         &  43 & 97.7\% & 0.992 \\
WanX         &  41 & 80.5\% & 0.952 \\
\bottomrule
\end{tabular}
\end{table}

\begin{table}[!t]
\centering
\caption{GenBuster-200K in-domain binary detection.}
\label{tab:gb200k}
\renewcommand{\arraystretch}{1.12}
\begin{tabular}{@{}lc@{}}
\toprule
\textbf{Metric} & \textbf{Value} \\
\midrule
ROC-AUC       & 0.986 \\
Balanced Acc. & 0.945 \\
Real recall   & 97.1\% \\
Fake recall   & 91.9\% \\
\bottomrule
\end{tabular}
\end{table}

\FloatBarrier
% ── B. GenBusterBench ─────────────────────────────────────────────────────────
\subsection{GenBusterBench: Benchmark Generalisation}
\label{subsec:genbench}

\begin{figure}[htbp]
\centering
\includegraphics[width=0.72\columnwidth]{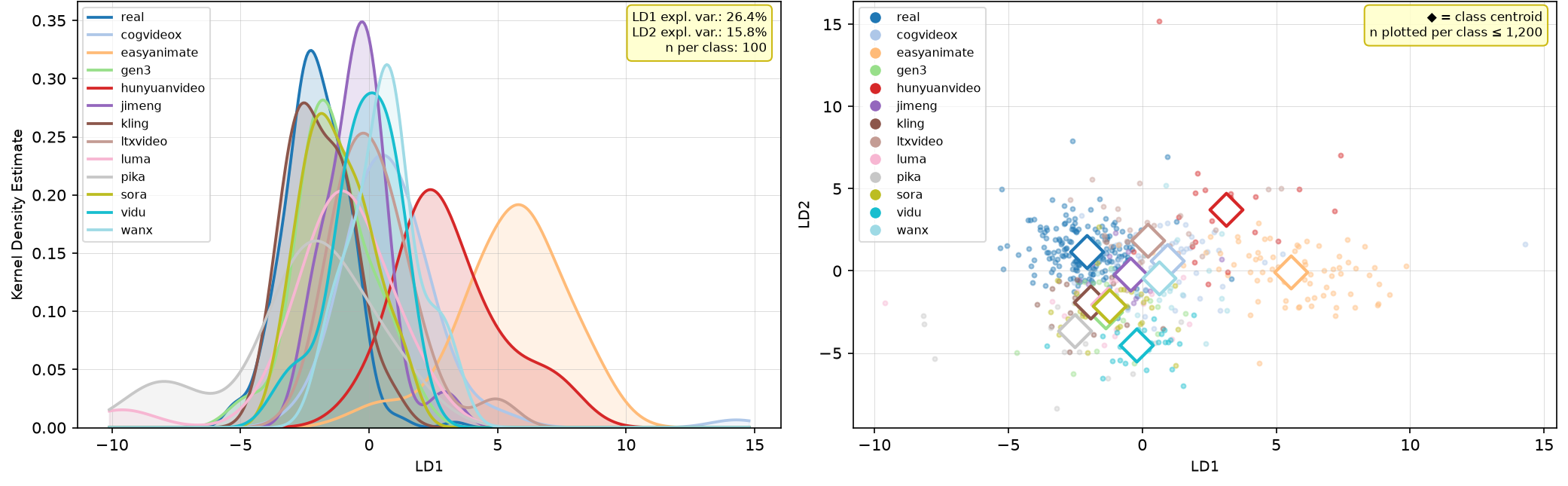}
\caption{LDA on GenBusterBench (all generators).  Multiple per-generator
clusters are visible in the scatter, each deviating from the real centroid
in a distinct direction---the geometric signature of generator-specific
artefact distributions.}
\label{fig:lda_genbench}
\end{figure}

\begin{figure}[htbp]
\centering
\includegraphics[width=0.72\columnwidth]{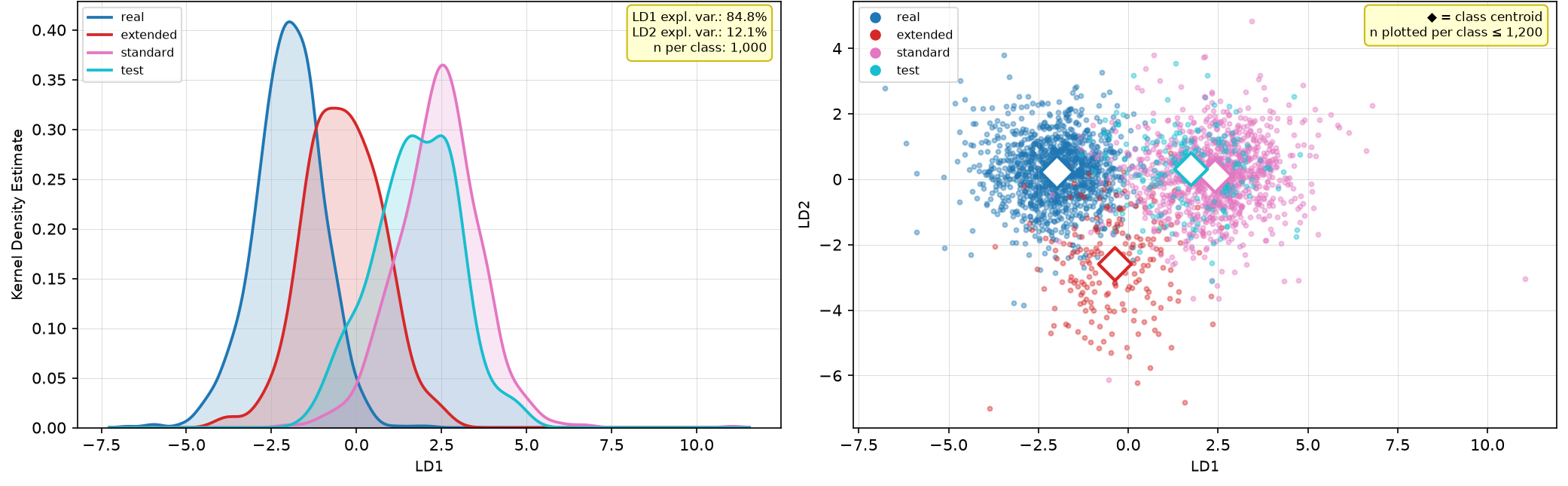}
\caption{LDA on GenBuster-200K features coloured by distribution group
(real, in-distribution fakes, benchmark fakes, test fakes).
In-distribution and benchmark fakes deviate from the real class in
\emph{different} LD directions.}
\label{fig:lda_gb200k_grouped}
\end{figure}

Table~\ref{tab:genbench_splits} summarises results across the three
GenBusterBench evaluation subsets.  Performance on the standard subset
(AUC 0.976) and extended subset (AUC 0.984) are comparable, indicating
that the feature captures a broadly shared artefact signature across
contemporary text-to-video generators regardless of whether a generator
family was represented in the training corpus.  Performance drops to
AUC 0.939 on the \emph{wild} subset of the latest-generation models, where
synthesis quality has advanced furthest and artefact magnitudes are
smallest.

\begin{table}[!t]
\centering
\caption{GenBusterBench results by evaluation tier.}
\label{tab:genbench_splits}
\renewcommand{\arraystretch}{1.12}
\setlength{\tabcolsep}{4pt}
\begin{tabular}{@{}lcccc@{}}
\toprule
\textbf{Tier} & \textbf{AUC} & \textbf{BalAcc} &
\textbf{Real Rec.} & \textbf{Fake Rec.} \\
\midrule
Standard            & 0.976 & 0.938 & 93.0\% & 94.5\% \\
Extended            & 0.984 & 0.950 & 93.0\% & 97.0\% \\
Wild (newest)       & 0.939 & 0.898 & 93.0\% & 86.7\% \\
\bottomrule
\end{tabular}
\end{table}

Fig.~\ref{fig:lda_gb200k_grouped} illuminates the underlying geometry.
The standard-subset and extended-subset fakes each form a cluster that
deviates from real in a different direction in the LDA space.  The MLP learns
a fixed decision boundary aligned to the artefact direction seen during
training; it cannot fully intercept artefacts that propagate in an orthogonal
direction.  This directional divergence is the structural reason why
performance degrades on the wild subset, independent of training-set size.

\begin{figure}[htbp]
\centering
\includegraphics[width=0.6\columnwidth]{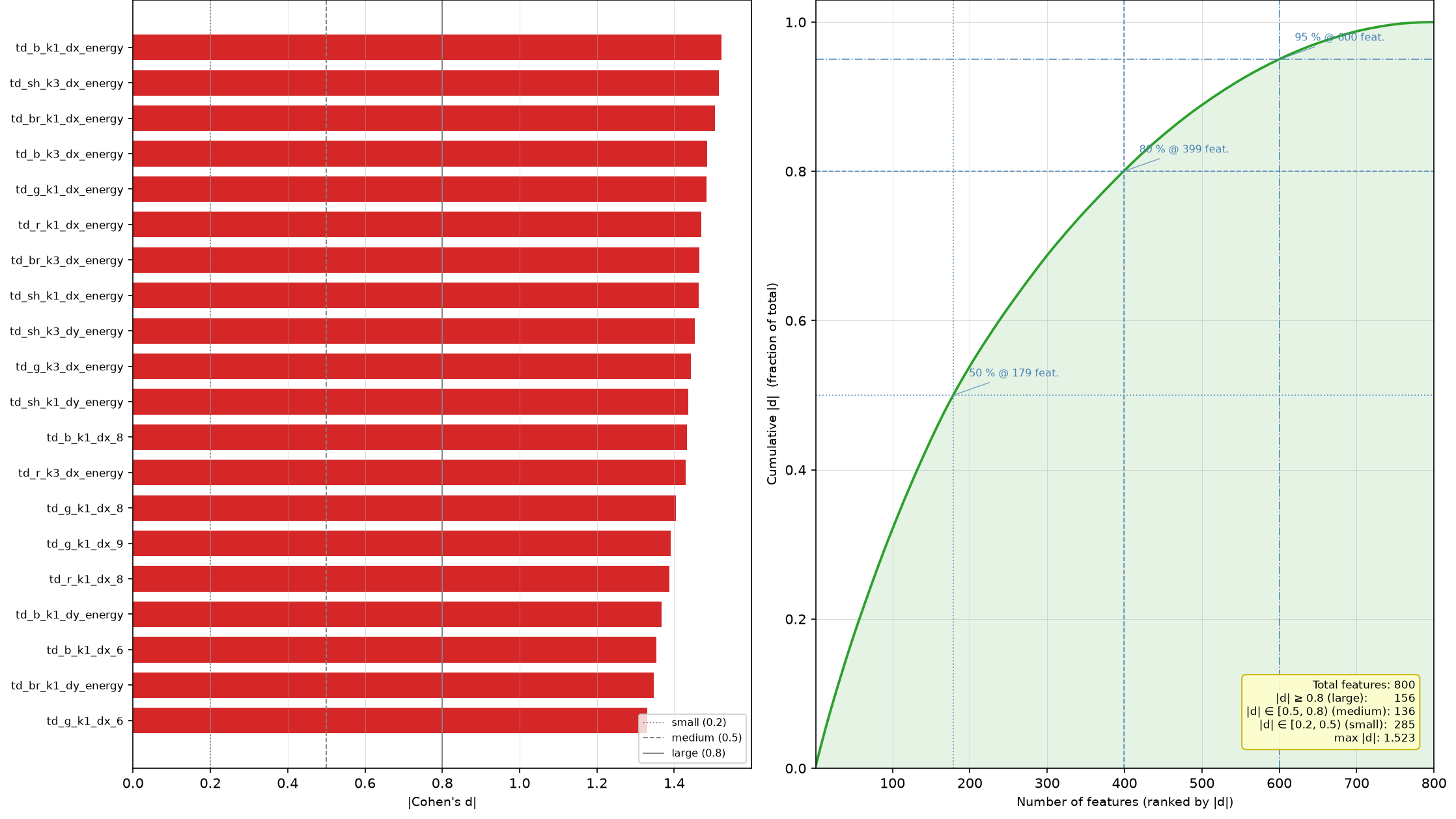}
\caption{Cohen's $d$ on GenBusterBench.  All top-20 features are
negative (red); effect sizes are uniformly large ($|d|\gtrsim0.8$),
with energy features dominating.}
\label{fig:cohen_genbench}
\end{figure}

Fig.~\ref{fig:cohen_genbench} shows the Cohen's $d$ analysis on
GenBusterBench.  All top-20 features show a negative sign (fake $<$ real
mean) and nearly all reach a large effect size, with $|d|$ values extending
above 1.3.  The energy features (\texttt{td\_*\_energy},
\texttt{sp\_*\_energy}) dominate the ranking, indicating that the primary
artefact signature on GenBusterBench is suppressed gradient
\emph{magnitude} rather than a redistribution of first-digit histogram
bins.  The cumulative power curve rises more slowly than for GB200K
(50\% ${\sim}174$ features, 80\% ${\sim}399$ features), reflecting the
greater diversity of generator families on this benchmark: no single cluster
of features captures the full discriminative signal, because each generator
contributes a slightly different flavour of gradient suppression.

\begin{figure}[htbp]
\centering
\includegraphics[width=0.3\columnwidth]{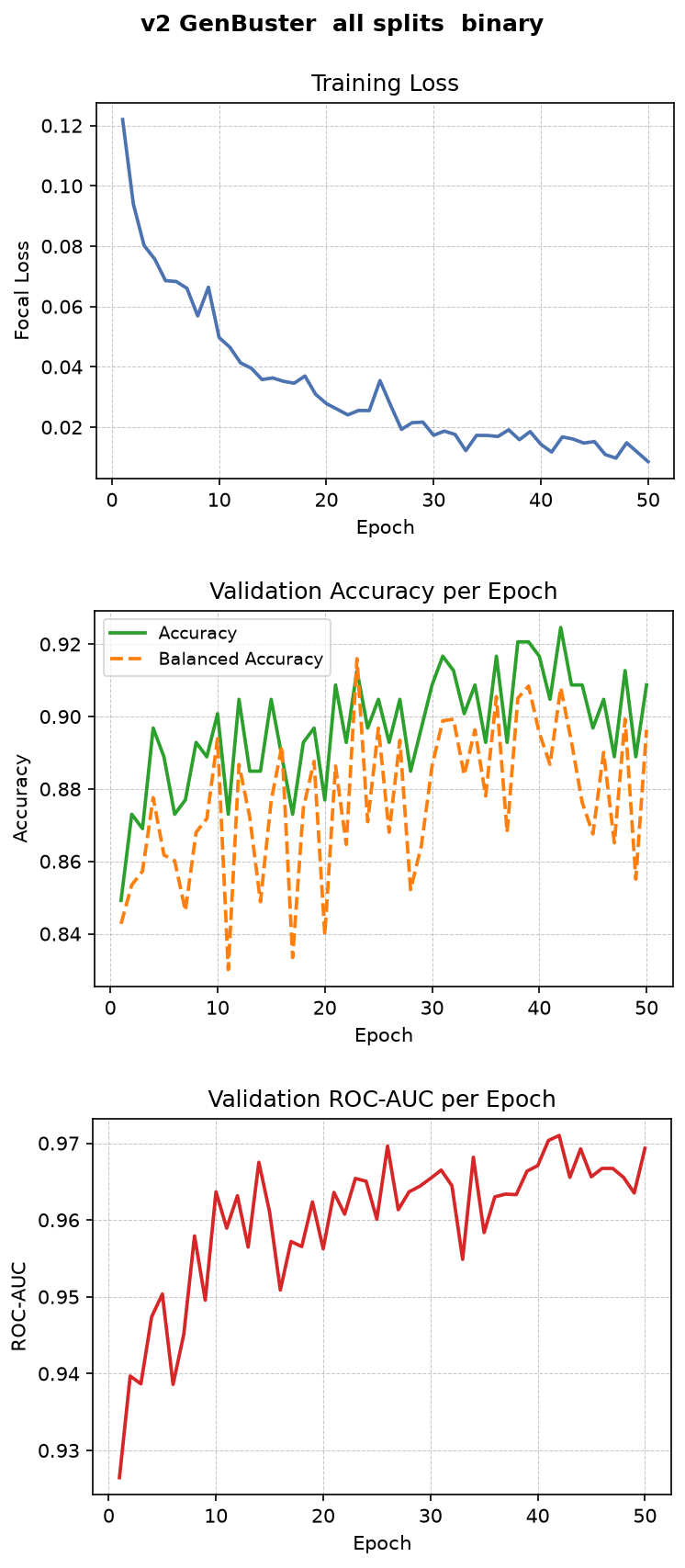}
\caption{Training curves for the GenBusterBench MLP (loss, balanced
accuracy, and ROC-AUC per epoch).}
\label{fig:genbench_curves}
\end{figure}

\begin{table}[!t]
\centering
\caption{GenVA zero-shot detection results (no fine-tuning).}
\label{tab:genva}
\renewcommand{\arraystretch}{1.12}
\begin{tabular}{@{}lrc@{}}
\toprule
\textbf{Generator} & $n$ & \textbf{Det.\ rate} \\
\midrule
Pika~v1        & 5{,}447 & 99.7\% \\
Sora           & 5{,}452 & 99.2\% \\
VideoCrafter-2 & 5{,}452 & 97.1\% \\
\midrule
\textbf{Overall} & \textbf{16{,}351} & AUC 0.993 / BalAcc 0.979 \\
\bottomrule
\end{tabular}
\end{table}

% ── C. GenVA ──────────────────────────────────────────────────────────────────
\subsection{GenVA: Zero-Shot Cross-Domain Generalisation}

\begin{figure}[htbp]
\centering
\includegraphics[width=0.72\columnwidth]{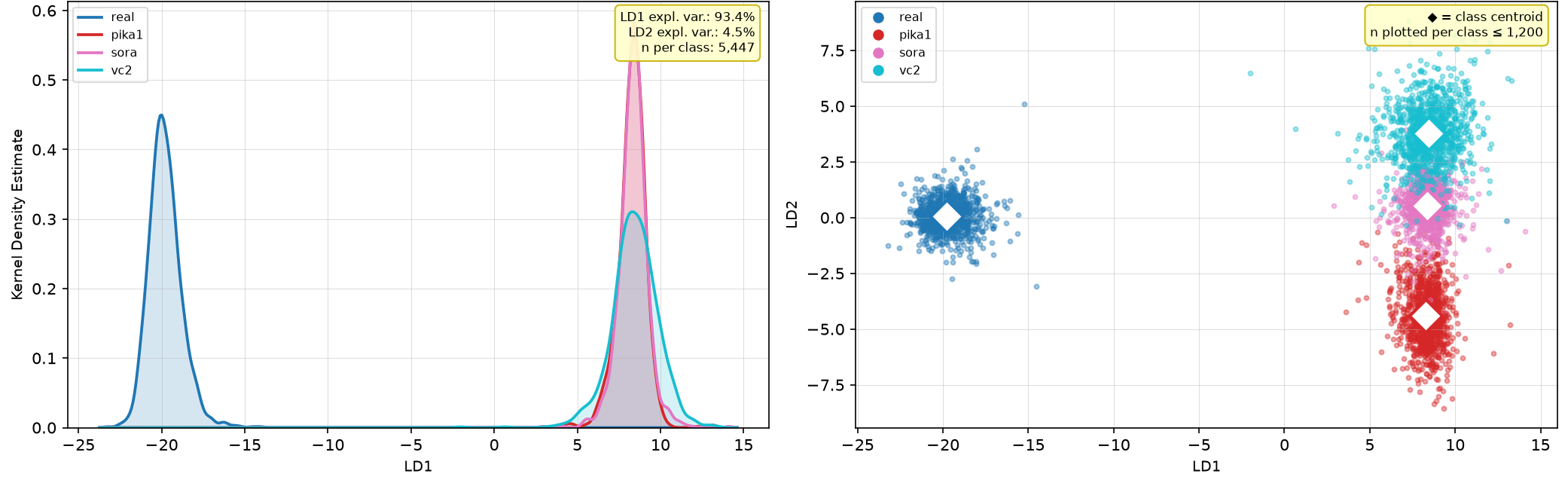}
\caption{LDA on GenVA fakes combined with real videos from the
GenBuster-200K test split.  All three GenVA generators form tight clusters
clearly separated from the real class, despite the model never having been
trained on GenVA.}
\label{fig:lda_genva}
\end{figure}

The GenVA experiment tests zero-shot cross-domain generalisation: the model
is trained exclusively on GB200K and evaluated on GenVA without fine-tuning.
Overall AUC reaches 0.993 and balanced accuracy 0.979 (Table~\ref{tab:genva});
Fig.~\ref{fig:lda_genva} shows the LDA projection.

It should be noted that Pika~(v1) and Sora are both present in the GB200K
training set, so their near-perfect detection rates (99.7\% and 99.2\%) are
partly expected---the model has already encountered artefact distributions
from those generator families and the cross-domain transfer is thus
partially in-distribution.  VideoCrafter-2 (VC2), which was never seen
during training, achieves 97.1\% detection, demonstrating genuine
cross-domain generalisation to an unseen synthesis architecture.  The
tight real cluster and well-separated fake generator clusters in the LDA
scatter confirm that text-to-video generators introduce consistent
gradient-statistic deviations that persist across model architectures and
remain detectable without retraining.

\begin{figure}[htbp]
\centering
\includegraphics[width=0.56\columnwidth]{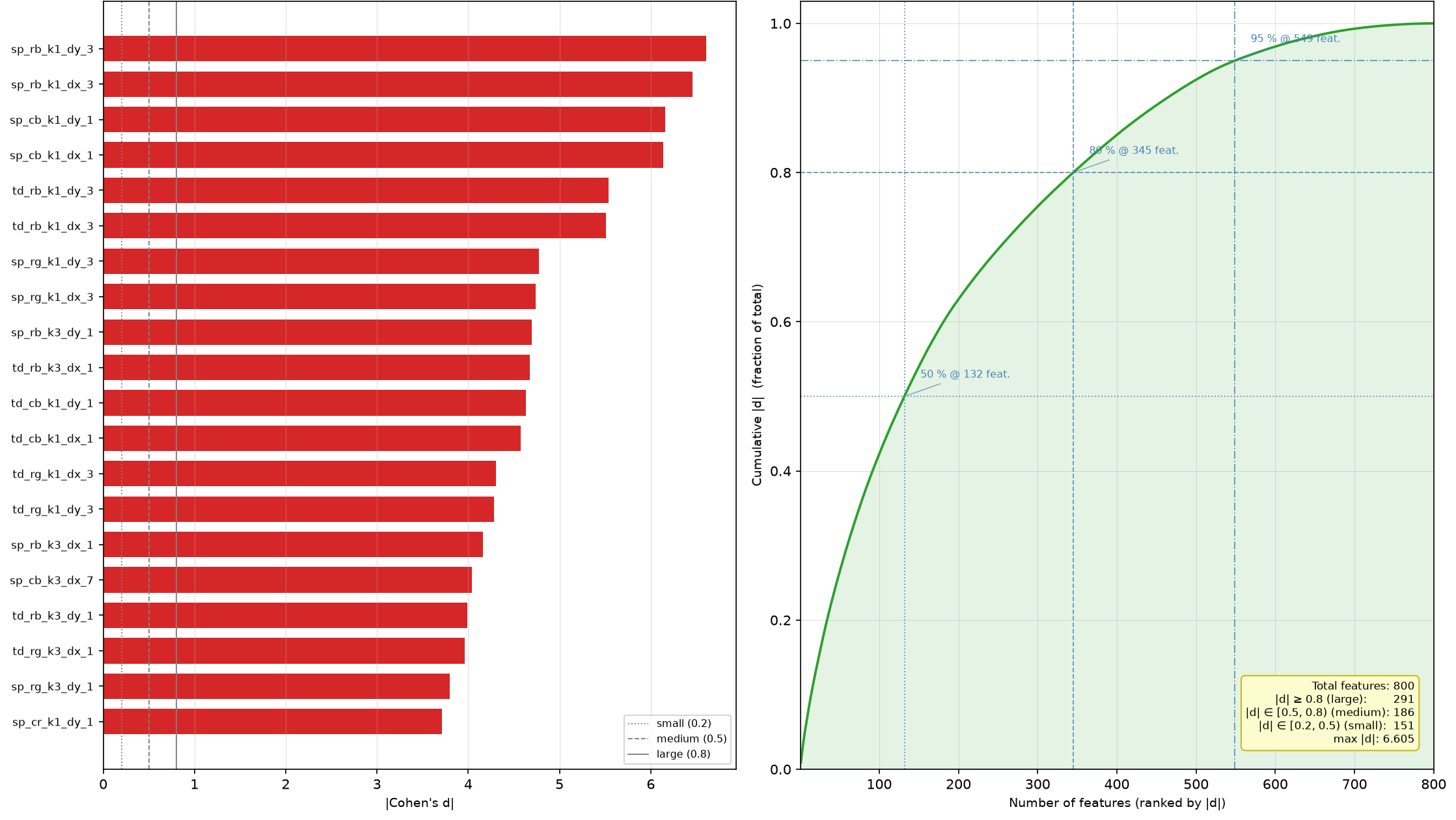}
\caption{Cohen's $d$ on GenVA.  Effect sizes are extreme ($|d|$ up to ${\sim}6$); the steep cumulative curve indicates the signal is concentrated in a small feature subset, consistent with near-perfect zero-shot detection.}
\label{fig:cohen_genva}
\end{figure}

Fig.~\ref{fig:cohen_genva} shows the Cohen's $d$ profile for GenVA.
Effect sizes are extreme---the top feature reaches $|d|\approx6$---and all
top features are negative (fake $<$ real), confirming that text-to-video
generators produce dramatically smoother gradient distributions than
authentic video.  The cumulative power curve rises steeply: 50\% of total
discriminative power is captured by ${\sim}132$ features and 80\% by
${\sim}291$ features.  This concentration of signal is consistent with the
near-perfect classification results: the MLP achieves high accuracy with
relatively few features because the effect sizes are so large.  The extreme
magnitude of the GenVA Cohen's $d$ values relative to GenBusterBench
reflects the fact that Sora, Pika, and VideoCrafter-2 produce outputs that
are more globally smooth---amplifying the systematic departure from Benford's
law across many feature channels simultaneously.

\FloatBarrier
% ── D. FaceForensics++ ────────────────────────────────────────────────────────
\subsection{FaceForensics++: Face-Swap Analysis}

\begin{figure}[htbp]
\centering
\includegraphics[width=0.72\columnwidth]{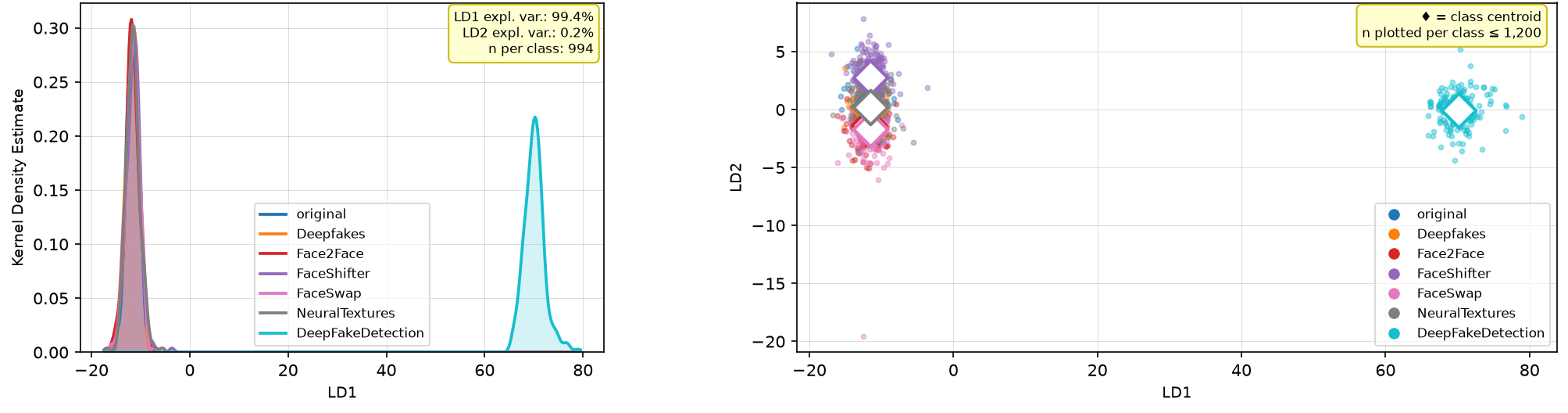}
\caption{LDA on FaceForensics++ restricted to fake classes only.  The five
forgery methods form distinct clusters, yet their joint separation from the
real class (cf.\ Fig.~\ref{fig:lda_pls_ff}) is limited---consistent with
the pixel-mass hypothesis: stitching-based forgeries transplant real pixels
rather than synthesizing new gradient distributions.}
\label{fig:ff_lda_fake}
\end{figure}

\begin{figure}[htbp]
\centering
\includegraphics[width=0.72\columnwidth]{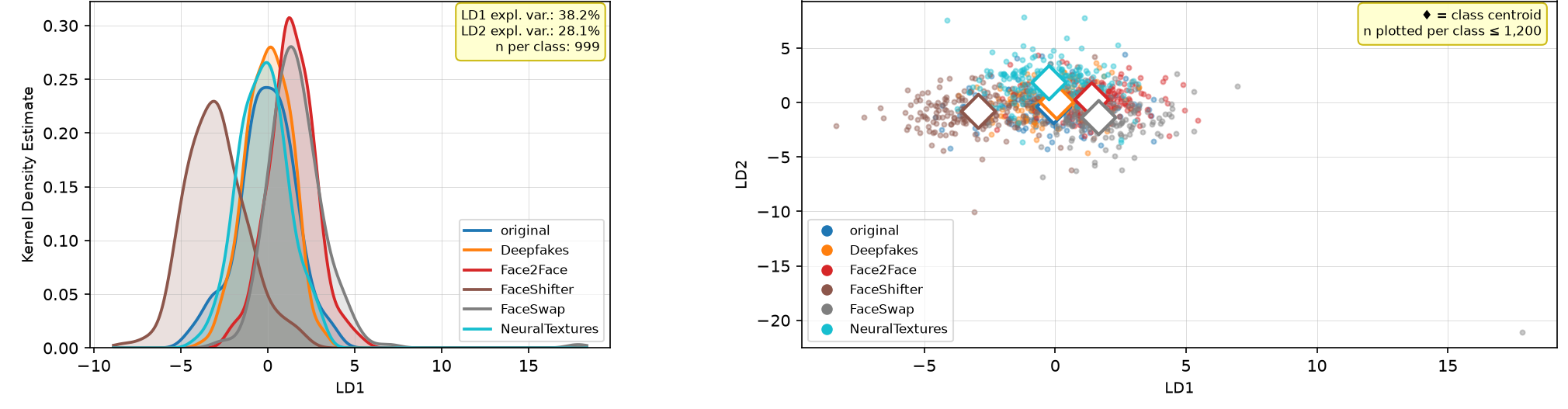}
\caption{LDA vs.\ PLS-DA on FaceForensics++ (six classes).  Rows from top:
LDA distribution / scatter, PLS-DA distribution / scatter, PLS2 distribution
and LD1-vs-PLS2 cross-scatter.  The marginal separation under both projections
(LDA BalAcc\,=\,0.521) confirms limited discriminability of face-swap
artefacts in the first-digit gradient feature space.}
\label{fig:lda_pls_ff}
\end{figure}

All FaceForensics++ experiments use the C23 (high-quality MPEG) compression
variant and a 1,600-dimensional feature: the standard 800-D video feature
concatenated with an additional 800-D feature computed on the face-cropped
region (background pixels zeroed), giving the classifier direct access to
the manipulated area.  The DeepFakeDetection class (994 clips) is withheld
from training as an eval-only class: Fig.~\ref{fig:ff_lda_fake} confirms
it would achieve near-100\% detection if trained on, as it is clearly
separable in the discriminant space.  It is excluded to prevent its
994 clips---the single largest subset in the corpus---from dominating the
Deepfakes direction, biasing the decision boundary, and producing an
artificially inflated headline accuracy; reporting it as eval-only yields
a more conservative and honest evaluation.

\begin{figure}[htbp]
\centering
\includegraphics[width=0.72\columnwidth]{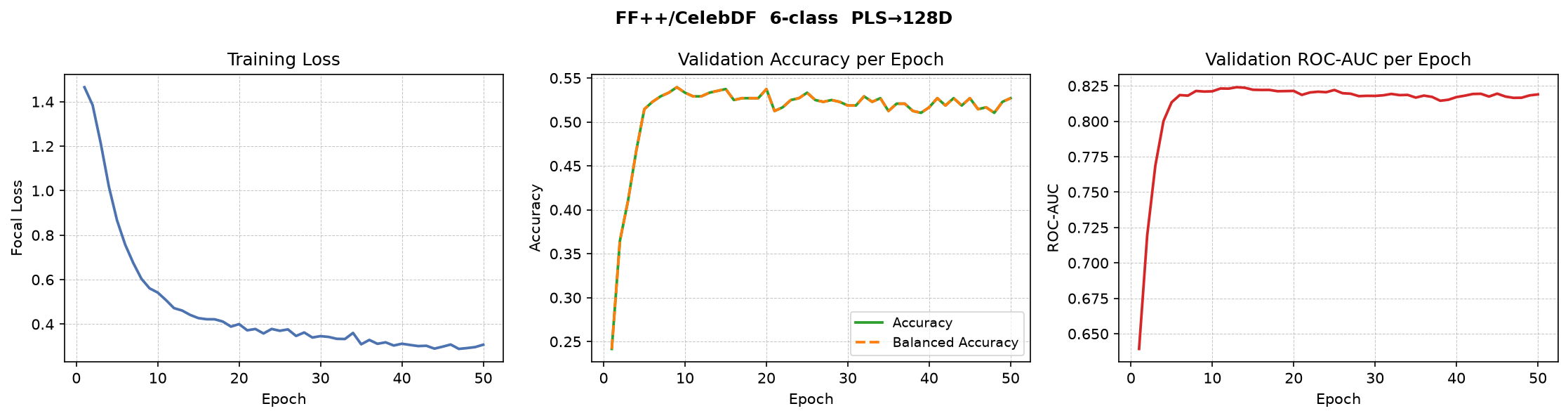}
\caption{Training curves for the FaceForensics++ PLS+MLP pipeline.}
\label{fig:ff_curves}
\end{figure}

The six-class classifier achieves balanced accuracy 0.524 and a collapsed
binary AUC of 0.717 (Table~\ref{tab:ff_binary}), substantially below the
text-to-video results.  The per-class recall ordering
(Table~\ref{tab:ff_perclass}) is: FaceShifter (77.5\%) $>$ NeuralTextures
(58.0\%) $>$ FaceSwap (56.0\%) $>$ Face2Face (50.5\%) $>$ Deepfakes
(43.0\%).  DeepFakeDetection, which was withheld from training, achieves
a detection rate of 79.4\% at inference, suggesting that the learned
representation transfers partially to unseen manipulation methods.
Training dynamics are shown in
Fig.~\ref{fig:ff_curves}.

\begin{table}[!t]
\centering
\caption{FaceForensics++ collapsed binary metrics (real vs.\ all fakes).}
\label{tab:ff_binary}
\renewcommand{\arraystretch}{1.12}
\begin{tabular}{@{}lc@{}}
\toprule
\textbf{Metric} & \textbf{Value} \\
\midrule
ROC-AUC (binary) & 0.717 \\
Balanced Acc.    & 0.596 \\
Real recall      & 29.5\% \\
Fake recall      & 89.6\% \\
\bottomrule
\end{tabular}
\end{table}

\begin{table}[!t]
\centering
\caption{FaceForensics++ per-class recall.}
\label{tab:ff_perclass}
\renewcommand{\arraystretch}{1.12}
\begin{tabular}{@{}lc@{}}
\toprule
\textbf{Class} & \textbf{Recall} \\
\midrule
Deepfakes              & 43.0\% \\
Face2Face              & 50.5\% \\
FaceShifter            & 77.5\% \\
FaceSwap               & 56.0\% \\
NeuralTextures         & 58.0\% \\
\midrule
DeepFakeDetection$^*$  & 79.4\% \\
\bottomrule
\multicolumn{2}{@{}l}{\footnotesize $^*$ Eval-only; withheld from training.}
\end{tabular}
\end{table}

\begin{figure}[htbp]
\centering
\includegraphics[width=0.56\columnwidth]{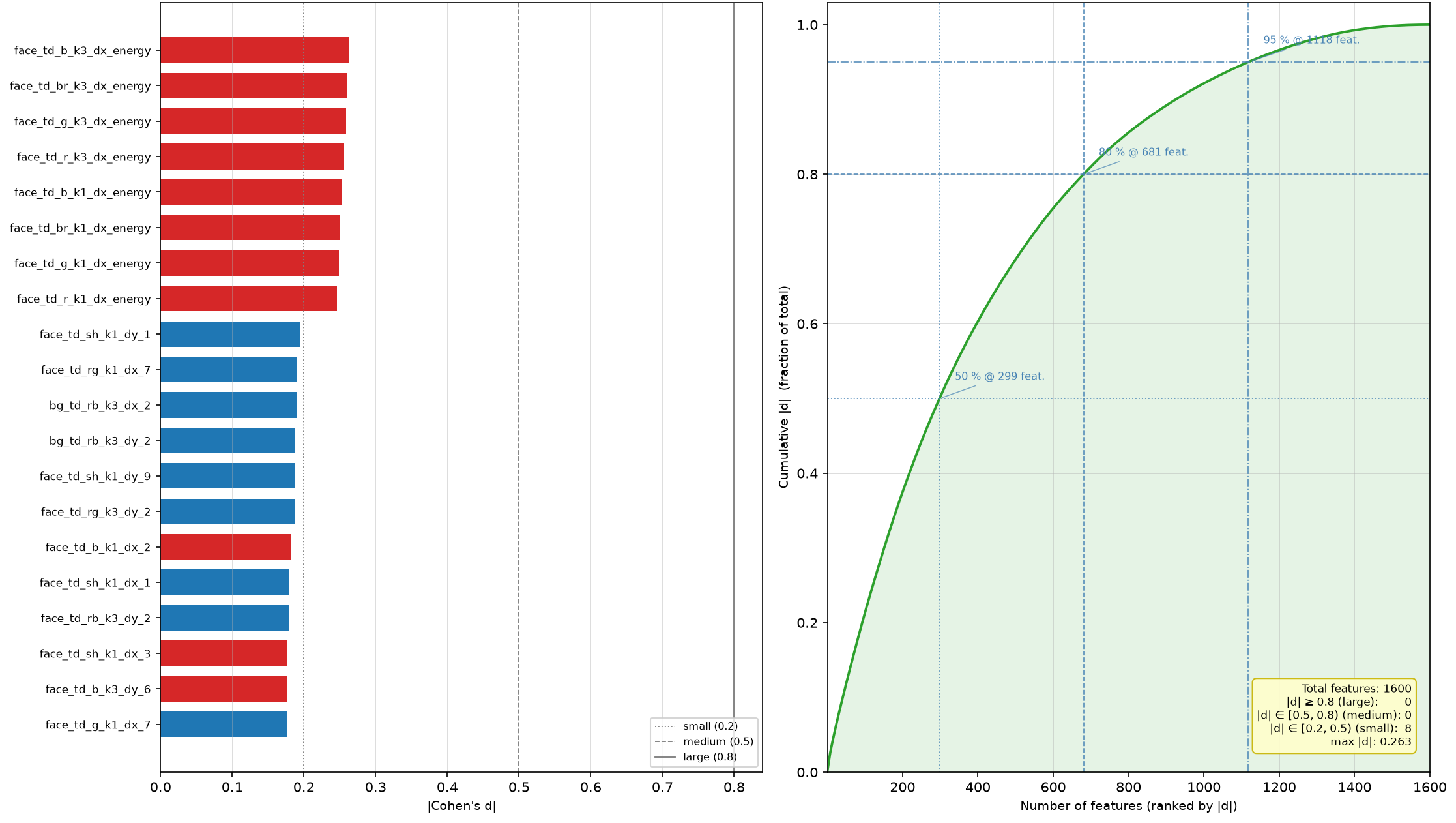}
\caption{Cohen's $d$ on FaceForensics++ (original vs.\ all manipulated pooled, 1,600-D face+background).  Effect sizes are uniformly small ($|d|<0.3$); the nearly linear cumulative curve is the quantitative signature of the pixel-mass limitation.}
\label{fig:cohen_ff}
\end{figure}

Fig.~\ref{fig:cohen_ff} provides the Cohen's $d$ profile for
FaceForensics++.  In contrast to the text-to-video datasets, the
maximum effect size $|d|$ barely exceeds 0.3, placing every feature well
below the conventional ``small'' threshold ($|d|=0.2$). The highest ranked
features are the temporal-derivative features which were extracted from the
(\texttt{face\_td\_*}), suggesting a possibility that the weak discriminative
signal is concentrated around the manipulated area and is primarily captured
by the temporal inconsistencies. 
The cumulative power curve is nearly linear (50\% at ${\sim}661$ features, 80\%
at ${\sim}961$ features out of 1,600), indicating that the discriminative
information is distributed across many features rather than being concentrated
in a small group that stands out. This diffuse signal is consistent with the
low LDA balanced accuracy and provides empirical support for the pixel-mass
hypothesis: when only a small facial region is manipulated and the remaining pixels
originate from genuine footage, the deviation from the natural-image statistics
is observed to be substantially weaker than that observed in fully synthetic videos.

The per-class recall ordering is not random: it tracks how much of the
frame each forgery type synthesizes.  Face-swap methods replace only the
face region; surrounding pixels are unmodified real footage.  The feature
vector accumulates first-digit gradient statistics across all sampled pixels
at stride~8 over the full frame; when the manipulated region occupies a
small fraction of total frame area, the synthetic signal is diluted by the
dominant unmodified background, suppressing the measurable Benford
deviation.  Crucially, stitching operations transplant real pixel patches
from a source identity---they do not synthesize new gradient
distributions---so the first-digit statistics of the composited region
remain close to natural-image statistics.  The additional 800-D face-crop
feature was included precisely to address this dilution problem; despite
directly exposing the classifier to the manipulated pixels, results remain
limited because the stitched content consists of real pixels from another
person's face and no new pixel patterns are introduced for the feature to
detect.

By contrast, modern text-to-video generators synthesize every pixel from a
learned prior, producing frame-wide deviations from natural gradient
distributions that our feature captures reliably.  The per-class recall
ordering is consistent with this \emph{pixel-mass hypothesis}: FaceShifter
and NeuralTextures are more recent forgeries that typically occupy a larger
blend region than earlier Deepfakes pipeline outputs, producing a larger
effective area of synthesized pixels and a stronger detectable first-digit
signal.  The low original-class recall (29.5\%) reflects the conservative
threshold calibrated to maximise fake recall; it is a tuning artefact
rather than a fundamental feature limitation.

\FloatBarrier
% ── E. CelebDF ─────────────────────────────────────────────────────────────────
\subsection{CelebDF: Cross-Dataset Face-Swap Check}
\label{subsec:celebdf}

CelebDF~\cite{li2020celebdf} provides a second, independently sourced
face-swap benchmark to check whether the FaceForensics++ finding above
generalises beyond a single forgery corpus.  The
standard 800-D video feature (no face-crop bank) is trained on an 80/20
stratified split (5,222 train, 1,306 test; 889 real vs.\ 5,639 Celeb-synthesis
fakes) using the same MLP architecture and focal-loss training recipe as the
other datasets, at a fixed 0.5 decision threshold (no validation-set threshold
calibration, unlike GB200K and GenBusterBench).  Figure~\ref{fig:celebdf_samples}
shows a representative real/fake pair; unlike the text-to-video generators
above, the Celeb-synthesis manipulation is confined to the face region and is
visually subtle.

\begin{figure}[htbp]
\centering
\includegraphics[width=0.42\columnwidth]{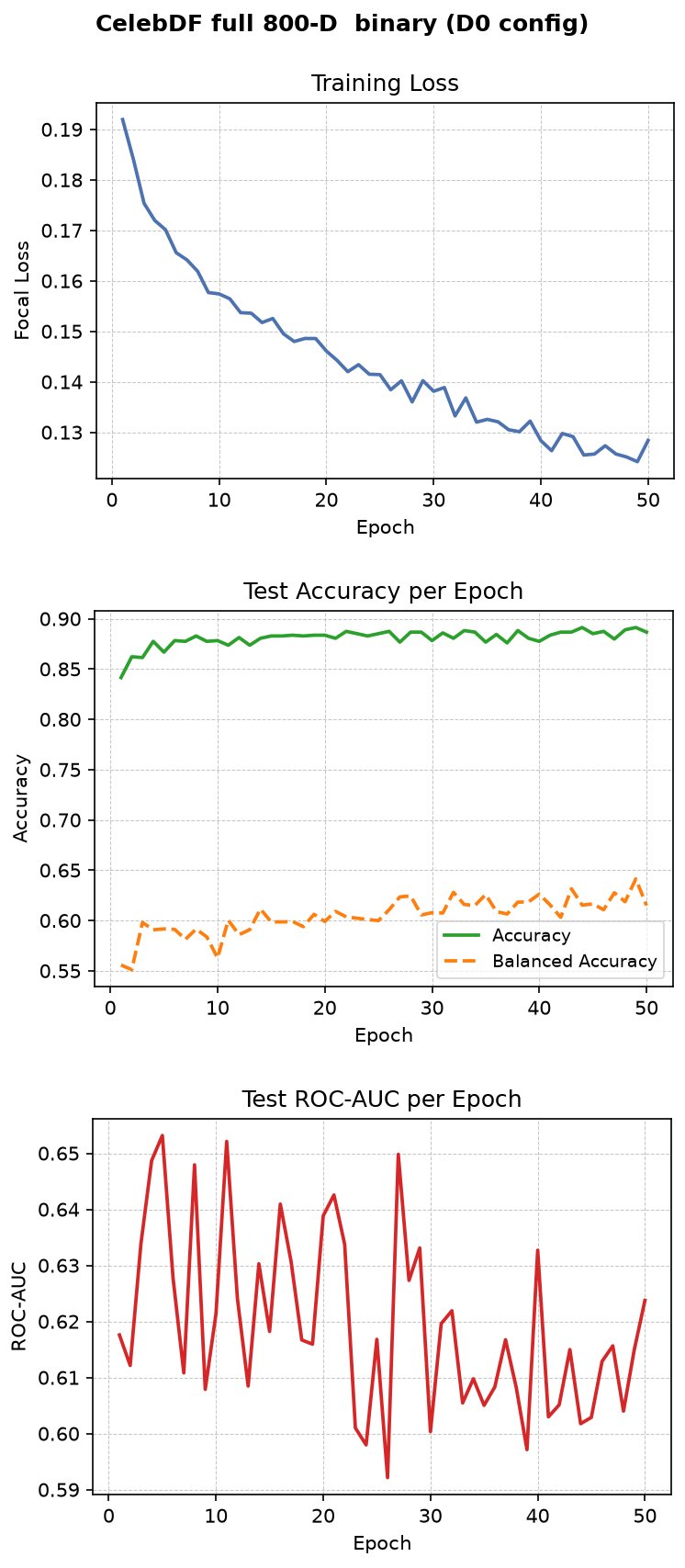}
\caption{CelebDF MLP training curves (loss, test accuracy/balanced accuracy,
and test ROC-AUC per epoch; full 800-D feature, 80/20 split).}
\label{fig:celebdf_curves}
\end{figure}

Table~\ref{tab:celebdf} summarises the result.  ROC-AUC (0.624) and balanced
accuracy (0.615) are well above chance but far below the 0.94--0.99 AUC range
achieved on the fully synthesized text-to-video benchmarks, and are close to
the FaceForensics++ collapsed-binary result (AUC 0.717).  The lopsided recall
split (fake 98.8\% vs.\ real 24.2\%) at the uncalibrated 0.5 threshold reflects
the same class imbalance and conservative-threshold effect discussed for
FaceForensics++, rather than a fundamentally different failure mode.  This
result is consistent with the \emph{pixel-mass hypothesis}
(Section~\ref{sec:conclusion}): CelebDF's Celeb-synthesis fakes are, like the
FaceForensics++ manipulations, face-swap forgeries that transplant a real face
region rather than synthesizing new pixels, so the detectable first-digit
gradient deviation is weak and diluted by the unmodified background, on a
dataset that shares no generator, encoding pipeline, or source footage with
FaceForensics++.

\begin{table}[!t]
\centering
\caption{CelebDF binary detection (full 800-D feature, 80/20 split).}
\label{tab:celebdf}
\renewcommand{\arraystretch}{1.12}
\begin{tabular}{@{}lc@{}}
\toprule
\textbf{Metric} & \textbf{Value} \\
\midrule
ROC-AUC       & 0.624 \\
Balanced Acc. & 0.615 \\
Real recall   & 24.2\% \\
Fake recall   & 98.8\% \\
\bottomrule
\end{tabular}
\end{table}

\FloatBarrier
% ── F. Benford's Law Deviation Analysis ────────────────────────────────────────
\subsection{Benford's Law Deviation Analysis}
\label{subsec:benford}

To directly characterise the statistical behaviour underlying the classifier's
discriminative signal, we compared the empirical first-significant-digit (FSD)
histogram of every (bank, channel, kernel size, direction) mode --- 80 modes in
total, spanning both the spatial and temporal-derivative banks across all ten
feature channels (Table~\ref{tab:channels}) --- against the theoretical Benford
curve $p(d)=\log_{10}(1+1/d)$~\cite{bonettini2020benford,chen2009firstdigit}.
The comparison is run on all five datasets used elsewhere in this paper
(GenBuster-200K, GenBusterBench, GenVA, FaceForensics++, and
CelebDF~\cite{li2020celebdf}).  Full per-mode overlays for all five datasets
are given in the Appendix (Fig.~\ref{fig:benford_appendix}).

Across every dataset, the extracted Sobel-gradient FSD histograms are
Benford-\emph{like}: probability mass is concentrated at the lower digits and
decreases monotonically toward digit~9, in qualitative agreement with the
theoretical curve.  However, every mode shows a systematic and reproducible
deviation from the logarithmic curve, appearing as localised excess mass at
specific digits rather than the smooth Benford decay --- most pronounced in
the raw-colour and channel-correlation channels ($r,g,b,\mathrm{rg},\mathrm{rb},
\mathrm{gb}$).  This deviation pattern is consistent across generators,
datasets, and video-encoding settings, which rules out sampling noise as the
explanation and instead indicates that it is an innate property of
derivative-domain frame statistics.  Critically, real and synthetic videos exhibit
this same deviation structure: the two classes are not separated by how
closely they adhere to Benford's law, so no video can be classified as
synthetic on that basis alone.  This confirms that the discriminative signal exploited by the classifier and
quantified by the Cohen's~$d$ analyses above is a shift in \emph{where}
probability mass sits away from a shared non-Benford baseline, not a
Benford-versus-non-Benford distinction.  The overall picture is a
stable, dataset-independent derivative-domain first-digit pattern that
departs from classical Benford behaviour in a structured and repeatable way,
which may itself be a useful statistical property for video forensics beyond
the classification task considered here.

\textbf{Quantifying stability independently of the Benford comparison.}
The consistency claim above compares each empirical histogram only to the
theoretical Benford curve; it does not by itself measure how similar the
empirical histograms are to \emph{each other}.  To quantify that separately,
for every one of the 80 modes we computed the Pearson correlation (shape
match) and the Jensen--Shannon (JS) distance --- the square root of the
Jensen--Shannon divergence, a bounded $[0,1]$ distributional distance with 0
indicating identical distributions --- between pairs of empirical mean FSD
histograms, then averaged over all 80 modes.  Two comparisons are made: (i)
\emph{across datasets}, holding class fixed (real-vs-real across four
independent real-video sources, and pooled fake-vs-pooled fake across all
five datasets); and (ii) \emph{across generators within a single dataset}
(fake-vs-fake, holding the dataset fixed).  Table~\ref{tab:stability}
summarises the result.

\begin{table}[!t]
\centering
\caption{Empirical-vs-empirical FSD histogram stability (not compared to the
theoretical Benford curve). Averaged over all 80 modes.}
\label{tab:stability}
\renewcommand{\arraystretch}{1.12}
\setlength{\tabcolsep}{4pt}
\begin{tabular}{@{}lrcc@{}}
\toprule
\textbf{Comparison} & \textbf{Pairs} & \textbf{mean $r$} & \textbf{mean JSD} \\
\midrule
Cross-dataset, real class          &  6 & 0.959 & 0.093 \\
Cross-dataset, pooled fake class   & 10 & 0.961 & 0.081 \\
\midrule
Cross-generator, GB200K (12 gens.)        & 66 & 0.993 & 0.038 \\
Cross-generator, GenBusterBench (12 gens.) & 66 & 0.994 & 0.038 \\
Cross-generator, GenVA (3 gens.)          &  3 & 0.987 & 0.059 \\
Cross-generator, FaceForensics++ (6 cls.) & 15 & 0.990 & 0.025 \\
Cross-generator, CelebDF (3 srcs.)        &  3 & 0.999 & 0.016 \\
\bottomrule
\end{tabular}
\end{table}

Correlation is uniformly high ($r=0.93$--$0.999$) and JS distance uniformly
low ($\leq0.12$ for any cross-dataset pair, $\leq0.06$ for any
cross-generator pair), confirming quantitatively that the FSD histogram
\emph{shape} is stable.  Cross-generator stability exceeds cross-dataset
stability in every dataset, consistent with the LDA and Cohen's~$d$ findings
above: different generators mainly shift feature \emph{magnitude} (captured
by Cohen's~$d$ and the per-generator LDA directions) rather than the
underlying histogram \emph{shape}, whereas genuinely different scene content
(general video vs.\ faces) is what moves the shape.  The residual
cross-dataset variation concentrates in the chrominance and
channel-correlation modes at the smallest (k=1) kernel size (JSD up to
$0.18$ for individual modes), while sharpness and base-colour channels at
k=3 are the most stable individual modes (JSD $\approx0.03$).

\section{Conclusion}
\label{sec:conclusion}

In this work we provide a Sobel first-digit-law based AI detection pipeline.
We examine the success and pitfalls of this type of detection. We note that
purely AI-generated datasets such as GenBuster-200K and GenVA produce high AUC
scores of 0.99 and 0.99 respectively, while datasets using face swaps and a
smaller number of generated pixels, with stitching of two videos, produce much
lower AUC scores of 0.717 and 0.624 on FF++\_C23 and CelebDF respectively. Our
hypothesis is that the deviation produced in the first-digit signal is directly
proportional to the number of pixels generated.

Apart from the detector, we believe our main contribution is to show
discriminatory signal in the first-digit distribution of non-scale-invariant
properties. When the rate of change is uniform (the fps and the filter size)
the first-digit distribution is incredibly stable and is shown to remain
relatively the same across different scenes. This shows the first-digit signal
to be more universal than previously thought.

For this research the main improvement will be to use other non-scale-invariant
image properties and test their discriminatory strength and stability. The use of
other forensic tools in addition to ours is a possible direction as well.

\noindent\textbf{Author Contributions}
Sidharth Shanu: Conceptualization, Methodology, Formal Analysis,
Validation, Visualization, Writing~-- Original Draft Preparation.
Gautam Kumar: Software, Data Curation, Investigation, Writing~--
Editing.
Tej Singh: Supervision, Methodology, Validation, Resources, Project
Administration, Writing~-- Review \& Editing.

\medskip
\noindent\textbf{Funding} Not Applicable.

\medskip
\noindent\textbf{Data Availability} The datasets analysed during the
current study are publicly available.  GenBuster-200K and
GenBusterBench are released by their respective authors.
GenVA~\cite{kang2025geneva} is publicly available.
FaceForensics++~\cite{rossler2019faceforensics} is available at
\url{https://github.com/ondyari/FaceForensics}.
CelebDF~\cite{li2020celebdf} is available at
\url{https://github.com/yuezunli/celeb-deepfakeforensics}.

\medskip
\noindent\textbf{Code Availability} The analysis code developed for
this study is available from the corresponding author on reasonable
request.

\section*{Declaration}

\noindent\textbf{Conflicts of interest} The authors declare no
competing interests.

\bibliographystyle{IEEEtran}
\bibliography{references}

\appendix
\label{sec:appendix}

\begin{figure}[!h]
\centering
\includegraphics[width=0.44\columnwidth]{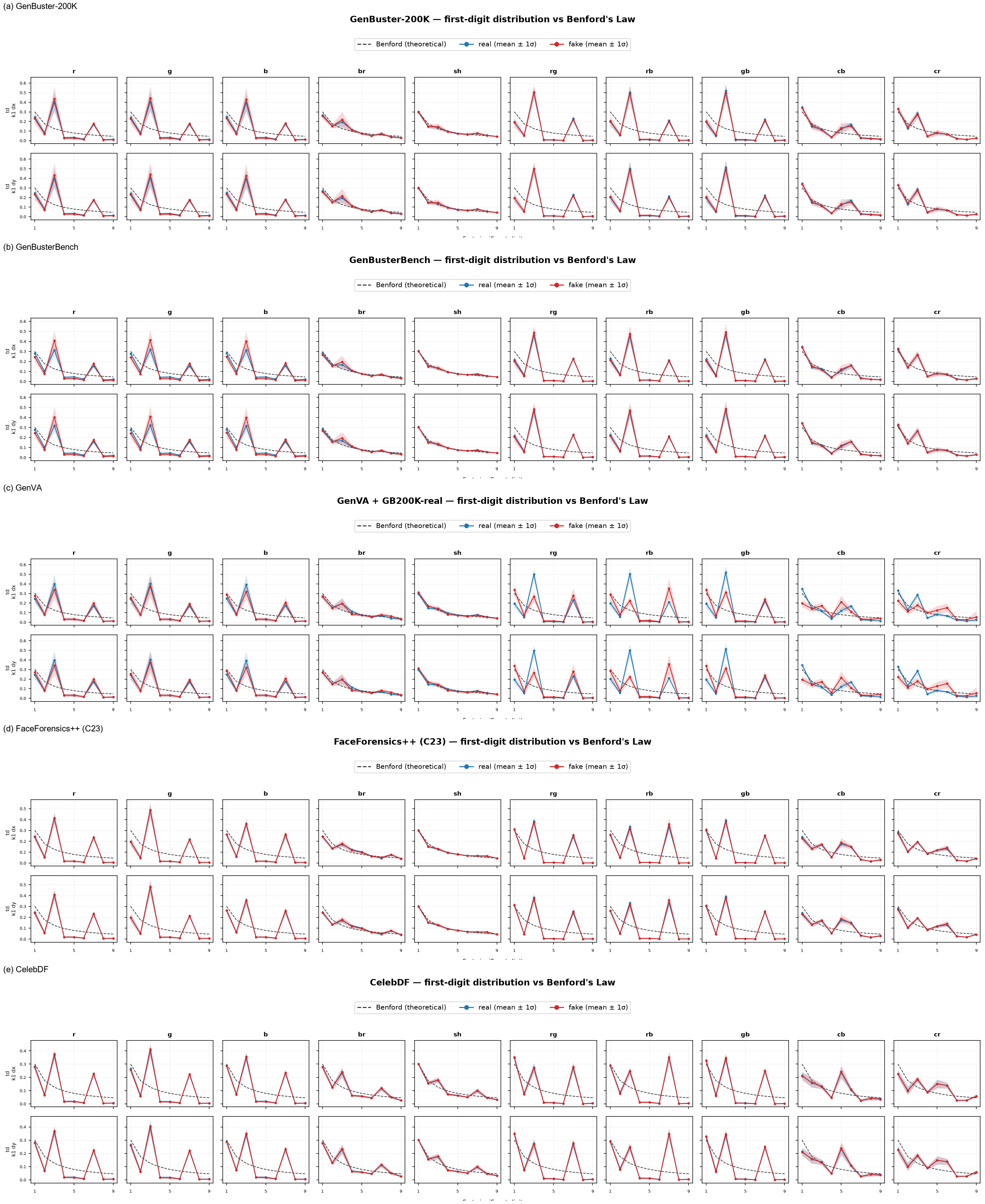}
\caption{Representative first-significant-digit distributions vs.\ the
theoretical Benford curve (temporal-derivative, $k{=}1$, both gradient
directions) for all five datasets: (a) GenBuster-200K, (b) GenBusterBench,
(c) GenVA (fakes combined with GenBuster-200K real videos), (d)
FaceForensics++ (C23, original vs.\ all manipulated classes pooled), (e)
CelebDF~\cite{li2020celebdf} (cf.\ Section~\ref{subsec:celebdf}
classification results).}
\label{fig:benford_appendix}
\end{figure}
\vspace{-3mm}

Fig.~\ref{fig:benford_appendix} gives a
representative sample of the per-mode overlays supporting the Benford's-law
deviation analysis of Section~\ref{subsec:benford}, for all five datasets.
Each panel-block shows two representative row-modes (temporal-derivative
bank, $k{=}1$, both gradient directions) across all ten feature channels of
Table~\ref{tab:channels}; the full 80-panel grids (two feature banks $\times$
two kernel sizes $\times$ two gradient directions $\times$ ten channels)
referenced in Section~\ref{subsec:benford} follow the same pattern.  Every
panel plots the theoretical Benford curve $p(d)=\log_{10}(1+1/d)$ (dashed
grey) against the empirical mean $\pm\,1$ standard deviation FSD histogram
for the real/original class (blue) and the fake/manipulated class (red),
computed over all videos of that class in the dataset.

\end{document}